\documentclass[runningheads]{llncs}

\usepackage{eccv}

\usepackage{eccvabbrv}

\usepackage{graphicx}
\usepackage{booktabs}
\usepackage{bm}
\usepackage{colortbl}
\usepackage{multirow}
\usepackage{arydshln}  
\usepackage{enumitem}
\usepackage{anyfontsize}  
\usepackage{algorithm}
\usepackage{algpseudocode}
\usepackage{amssymb}   
\usepackage{setspace}  

\usepackage[accsupp]{axessibility}  

\usepackage[pagebackref,breaklinks,colorlinks,citecolor=eccvblue]{hyperref}

\usepackage{orcidlink}

\begin{document}

\title{Unified Text-Image-to-Video Generation: \\ A Training-Free Approach to Flexible Visual Conditioning} 

\titlerunning{Unified Text-Image-to-Video Generation}

\author{Bolin Lai\inst{1} \quad
Sangmin Lee\inst{3} \quad
Xu Cao\inst{2} \quad
Xiang Li\inst{2} \quad
James M. Rehg\inst{2} \\
$^1$Georgia Tech \quad $^2$UIUC \quad $^3$Korea University \\
{\small\tt bolin.lai@gatech.edu \  sangmin-lee@korea.ac.kr \ \{xucao2,xiangl12,jrehg\}@illinois.edu} \\
{\small \textbf{Project Page: \textcolor{NavyBlue}{\url{https://bolinlai.github.io/projects/FlexTI2V/}}}}
}

\authorrunning{Lai et al.}

\institute{}

\maketitle

\begin{figure}
  \centering
  \vspace{-0.7cm}
  \includegraphics[width=\linewidth]{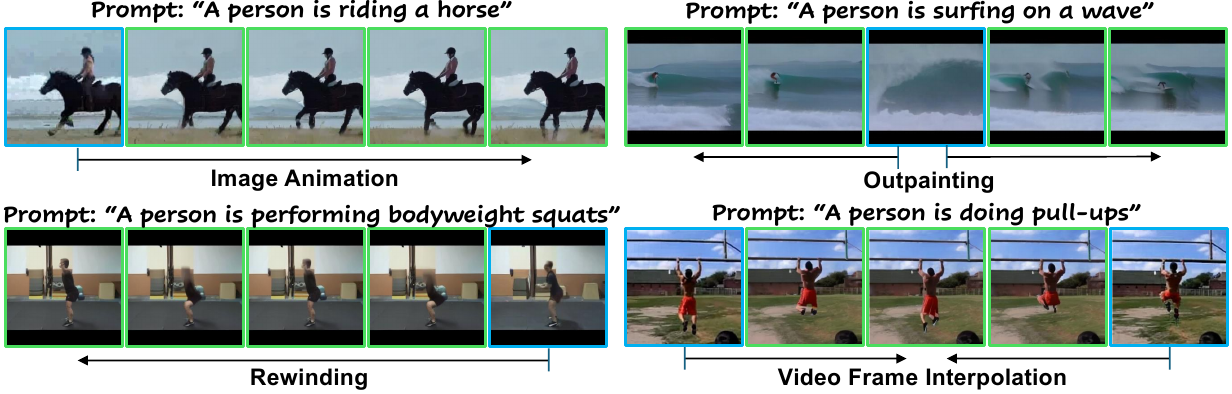}
  \vspace{-0.5cm}
  \captionof{figure}{We propose \textbf{FlexTI2V}, a novel training-free approach that can add flexible image conditioning to off-the-shelf text-to-video foundation models. We are able to frame an \textit{arbitrary number} of images at \textit{arbitrary positions} in the synthetic video with vivid motion and smooth transitions. In this figure, images with \textcolor{NavyBlue}{blue} edges are condition images, and images with \textcolor{ForestGreen}{green} edges are generated video frames.}
  \vspace{-0.5cm}
  \label{fig:teaser}
\end{figure}

\begin{abstract}
  Text-image-to-video (TI2V) generation is a critical problem for controllable video generation using both semantic and visual conditions. Most existing methods typically add visual conditions to text-to-video (T2V) foundation models by finetuning, which is costly in resources and only limited to a few pre-defined conditioning settings. To tackle these constraints, we introduce a unified formulation for TI2V generation with flexible visual conditioning. Furthermore, we propose an innovative training-free approach, dubbed \textbf{FlexTI2V}, that can condition T2V foundation models on an arbitrary amount of images at arbitrary positions. Specifically, we firstly invert the condition images to noisy representation in a latent space. Then, in the denoising process of T2V models, our method uses a novel random patch swapping strategy to incorporate visual features into video representations through local image patches. To balance creativity and fidelity, we use a dynamic control mechanism to adjust the strength of visual conditioning to each video frame. Extensive experiments validate that our method surpasses previous training-free image conditioning methods by a notable margin. Our method can also generalize to both UNet-based and transformer-based architectures across various image conditioning settings.
\end{abstract}

\section{Introduction}
\label{sec:intro}

In recent years, diffusion models \cite{ho2020denoising} have shown excellent performance in text-to-video (T2V) generation \cite{ho2022video,he2022latent,singer2023makeavideo,hongcog2023video,kong2024hunyuanvideo,blattmann2023align,wang2025wan,yang2025cogvideox,wang2023modelscope}, serving as video foundation models for various downstream video generation tasks. Given the vagueness and ambiguity of language \cite{lai2025unleashing,lai2024lego}, recent work has investigated synthesizing videos conditioned on RGB images for precise controllability on the appearance of people, objects and background \cite{blattmann2023stable,li2024generative,chen2024seine}, which is as known as image-to-video (I2V) generation. In these models, RGB images can be used as conditions alone \cite{blattmann2023stable,li2024generative,wang2022latent,mahapatra2022controllable}, or incorporated with textual prompts for accurate guidance \cite{zhang2023i2vgen,guo2023animatediff,xing2024dynamicrafter,huang2025step}. For clarity, we explicitly refer to the latter as text-image-to-video (TI2V) generation following \cite{huang2025step,ni2024ti2v}.

\begin{figure}[t]
  \centering
  \includegraphics[width=\linewidth]{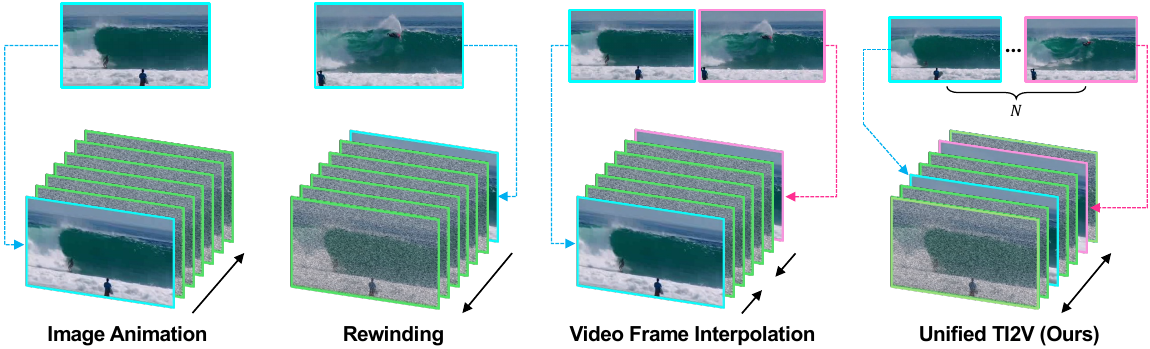}
  \caption{\textbf{Comparison with classic TI2V tasks.} Our task requires video generation conditioned on any number of images at any positions in the output video, which unifies existing classic TI2V tasks. The images with \textcolor{NavyBlue}{blue} and \textcolor{Lavender}{pink} edges are condition images, and images with \textcolor{ForestGreen}{green} edges are generated video frames.}
  \label{fig:i2v_tasks}
\end{figure}

To include extra image conditioning in video generation, the most widely-used approach is finetuning pretrained T2V model with paired texts, images and videos. However, the prohibitive demands of high-quality data and computing resources, and the risks of involving unexpected bias from the data during finetuning impede the development of TI2V models. Moreover, as shown in Fig. \ref{fig:i2v_tasks}, off-the-shelf TI2V models can only apply the image conditioning setting pre-defined during training (\textit{e.g.}, image animation \cite{guo2023animatediff,zhang2024pia,chen2024livephoto} and video frame interpolation \cite{sim2021xvfi,reda2022film}), lacking sufficient flexibility in the number and positions of condition images. Any changes in the conditioning setting require re-training of the entire model. To avoid the computing overhead and undesired bias of finetuning, training-free image conditioning approaches are promising solutions to incorporate flexible visual control into off-the-shelf T2V models.

Prior to our work, recent studies have been conducted on training-free controllable video generation with various conditions \cite{kara2024rave,khachatryan2023text2video,zhou2024storydiffusion,hou2025training,liang2025looking,liang2025movie}. For training-free image conditioning, the previous methods either synthesize only one video frame in one denoising pass, making it inefficient in inference \cite{ni2024ti2v}, or fill in the missing video frames between two bounded images solely based on visual cues, losing semantic control of generated frames in between \cite{feng2024explorative}. Similar to finetuned TI2V models, they are also only capable of following pre-defined conditioning settings, and still short of flexibility for image conditioning. These inevitable deficiencies make the existing training-free methods suboptimal for TI2V generation problem, thus demanding more efforts and novel solutions.

In this work, we formally define a generic TI2V generation problem with high flexibility in the amounts and positions of condition images (see Figs. \ref{fig:teaser}, \ref{fig:i2v_tasks}), unifying the existing TI2V tasks such as image animation, rewinding and video frame interpolation. To address aforementioned issues, we introduce \textbf{\textit{FlexTI2V}}, an innovative flexible training-free approach that can condition an off-the-shelf text-to-video diffusion model on an arbitrary number of images at arbitrary positions in the synthetic video. Specifically, we first reverse the images to noisy representations at each denoising step. To harmonize the difference in denoising paths of T2V models and image conditioning, we propose a novel \textit{random patch swapping} method to incorporate visual features of input images into each video frame, by swapping a random portion of patches between video frames and condition images. We also use a dynamic adjustment strategy to control the percentage of swapped patches and denoising steps for each frame to prevent frozen videos. The extensive experiments suggest that our approach outperforms the previous training-free image conditioning methods for video diffusion models. Overall, our main contributions are summarized as follows:

\begin{itemize}[itemsep=0.5ex, parsep=0pt, topsep=0pt, leftmargin=0.7cm]
    \item[$\bullet$]  We formally define a generic text-image-to-video generation problem, unifying the various settings of existing TI2V tasks.
    
    \item[$\bullet$] We introduce FlexTI2V, a novel training-free approach to incorporate flexible image conditioning into pretrained text-to-video diffusion models. The key innovation is a random patch swapping method to incorporate visual guidance into the denoising process in a plug-and-play manner.

    \item[$\bullet$] The experiments indicate that our approach notably surpasses the existing training-free image conditioning methods. Our method also generalizes seamlessly to both UNet and transformer architectures.
\end{itemize}

\section{Related Work}

\textbf{Video Generation Conditioned on Texts and Images}\quad Text-to-video (T2V) generation has gained impressive advancement since the emergence of diffusion models \cite{ho2022video,he2022latent,hong2022cogvideo,blattmann2023align,kong2024hunyuanvideo,yang2025cogvideox,wang2025wan,wang2023modelscope}. Given the ambiguity of language in describing visual details, RGB images are exploited as the condition for video generation. Image conditioning can be used alone for image-to-video (I2V) generation \cite{blattmann2021understanding,dorkenwald2021stochastic,wang2022latent,mahapatra2022controllable,blattmann2023stable,li2024generative}, or integrated with textual prompts for text-image-to-video (TI2V) generation \cite{ni2023conditional,zhang2023i2vgen,jiang2024videobooth,chen2024livephoto,wang2025dreamvideo,huang2025step}. Here, we mainly discuss previous TI2V methods. Hu et al. \cite{hu2022make} propose TI2V task for the first time, and develop a transformer-based architecture to achieve decent controllability and diversity in generated videos. Guo et al. \cite{guo2023animatediff} develop a practical framework, AnimateDiff, to train a model-agnostic plug-and-play motion module that can turn any personalized text-to-image diffusion model into an image animator. Xing et al. \cite{xing2024dynamicrafter} introduce a novel model -- DynamiCrafter, which can animate both human bodies and objects. Fu et al. \cite{fu2023tell} propose a unified video completion method that can tackle image animation, rewinding and infilling using a single model. Most prior methods require training some layers or the entire model to incorporate image conditioning or motion over time. As a result, existing models can only handle TI2V tasks pre-defined in training, such as animation and infilling. In this paper, we introduce a generic TI2V problem that unifies previous TI2V tasks. Our proposed method is able to condition T2V diffusion models on an arbitrary number of images at arbitrary positions of the output videos without finetuning.

\noindent\textbf{Video Frame Interpolation}\quad Video frame interpolation has been extensively studied in recent years, with the model architectures spanning across convolutional networks \cite{niklaus2017video,liu2020enhanced,reda2022film}, transformers \cite{fu2023tell} and diffusion models \cite{feng2024explorative,github2024cogvideox,github2025wan}. Liu et al. \cite{liu2017video} propose a convolutional network to synthesize video frames by flowing pixel values from bounded images. Xu et al. \cite{xu2019quadratic} point out that most interpolation methods assume a uniform motion between two consecutive images, thus resulting in oversimplified linear models for interpolation. To generate complex motion flows, They propose a quadratic interpolation method to model acceleration information in videos. Previous work mainly aims at finding a plausible path of transition from the first video frame to the last one. How to go beyond the two bounded images remains understudied. In this work, we work on the generic TI2V problem including both classic video frame interpolation and frame synthesis beyond the bound of start and end images.

\noindent\textbf{Training-free Methods for Diffusion Foundation Models}\quad The high computing overhead and large amounts of training data of finetune the foundation models are inevitable bottlenecks for downstream tasks. Recently, the development of training-free approaches provide an excellent trade-off between performance and cost \cite{hertz2022prompt,mokady2023null,tumanyan2023plug,liu2024towards,khachatryan2023text2video,kara2024rave,ni2024ti2v,titov2024guide,feng2024explorative,zhou2024storydiffusion,hou2025training,liang2025looking,liang2025movie,hsiao2025tf}. Hertz et al. \cite{hertz2022prompt} introduce prompt-to-prompt method to edit synthetic images by manipulating self-attention maps, which is further extended to real photo editing using null-text inversion \cite{mokady2023null}. Kara et al. \cite{kara2024rave} adapt a pretrained text-guided image editing model for video editing tasks by random noise shuffling. In terms of image-conditioned video generation, the work most related to our method is TI2V-Zero \cite{ni2024ti2v}. They duplicate the condition image and synthesize only one video frame in a single denoising process, which enforces the pretrained temporal layer to leverage visual features from the image. However, this method suffers from slow inference speed and compounding error. Feng et al. \cite{feng2024explorative} convert an image-to-video model to a bounded-image interpolation model by merging the two separate denoising paths conditioned on the bounded images. Hou et al. \cite{hou2025training} propose to map a single image to 3D point cloud and then synthesize new videos from other viewpoints. Existing training-free TI2V methods usually focus on a specific problem setting. We make the first attempt to develop a training-free method for various image conditioning settings, with high flexibility in image amounts and positions.

\section{Preliminaries}

\textbf{Latent Diffusion Models}\quad Denoising Diffusion Probabilistic Models (DDPM) \cite{ho2020denoising} are initially proposed in RGB space, which is computationally expensive because of the high resolution. As a solution, latent diffusion models (LDM) \cite{rombach2022high} project RGB images or videos into a low-dimensional latent space with a visual encoder $\mathcal{E}$. The denoising process is conducted in the latent space to generate the latent representation of desired images or videos, which are decoded to RGB space by a decoder $\mathcal{D}$. To achieve controllable generation, the conditions are also embedded into the same space, and typically merged into the model through cross-attention layers.

\noindent\textbf{Denoising Diffusion Implicit Models (DDIM)}\quad DDIM \cite{song2020denoising} basically include two stages -- diffusion stage and denoising stage. Given a clean representation $\bm{z}_0$ in the latent space, Gaussian noise is added to the $\bm{z}_0$ step by step in the diffusion stage. The single‑step sample $\bm{z}_t$ at step $t$ also follows a Gaussian distribution, $\bm{z}_t\sim\mathcal{N}(\sqrt{\bar{\alpha}_t}\;\bm{z}_0,\ (1-\bar{\alpha}_t)\bm{I})$. The closed-form equation is
\begin{equation}\label{eq:diffusion}
    \bm{z}_t = \sqrt{\bar{\alpha}_t}\cdot\bm{z}_0 + \sqrt{1-\bar{\alpha}_t}\cdot\bm{\epsilon}, \quad \bm{\epsilon}\sim\mathcal{N}(\bm{0},\bm{I}),
\end{equation}
where $\bar{\alpha}_t=\prod_{s=1}^t\alpha_s$ and $\alpha_t=1-\beta_t$. $\beta_t$ is the noise schedule. In denoising stage, DDPM build the denoising process as a Markov chain, and estimates the distribution of $\bm{z}_{t-1}$ from $\bm{z}_t$. To accelerate sampling, DDIM define a more general non-Markovian process to estimate the Gaussian distribution of $\bm{z}_{t-1}$ conditioned on $\bm{z}_t$ and $\bm{z}_0$, thus reducing sampling steps significantly. After substitution of $\bm{z}_0$ with an expression of $\bm{z}_t$, DDIM sampling is formulated as
\begin{equation}\label{eq:ddim} 
  \bm{z}_{t-1} = \frac{\sqrt{\bar{\alpha}_{t-1}}}{\sqrt{\bar{\alpha}_t}} \left(\bm{z}_t-\sqrt{1-\bar{\alpha}_t}\cdot\bm{\varepsilon}_\theta(\bm{z}_t,t)\right) + \sqrt{1-\bar{\alpha}_{t-1}-\sigma_t^2}\cdot\bm{\varepsilon}_\theta(\bm{z}_t,t) + \sigma_t\cdot\bm{\epsilon}'
\end{equation}
where $\bm{\epsilon}'\sim\mathcal{N}(\bm{0},\bm{I})$. $\bm{\varepsilon}_\theta$ is a neural network trained to predict the noise added to $\bm{z}_t$. $\sigma_t$ is a parameter that controls the stochasticity of the sampling process. Setting $\sigma_t=0$ yields a deterministic path that can reduce sampling steps with minor loss of generation quality \cite{song2020denoising}. Then Eq. \ref{eq:ddim} can be simplified to a function of $\bm{z}_t$ and $t$, \textit{i.e.}, $\bm{z}_{t-1}=\bm{\mu}_\theta(\bm{z}_t,t)$.

\begin{figure}[t]
  \centering
  \includegraphics[width=\linewidth]{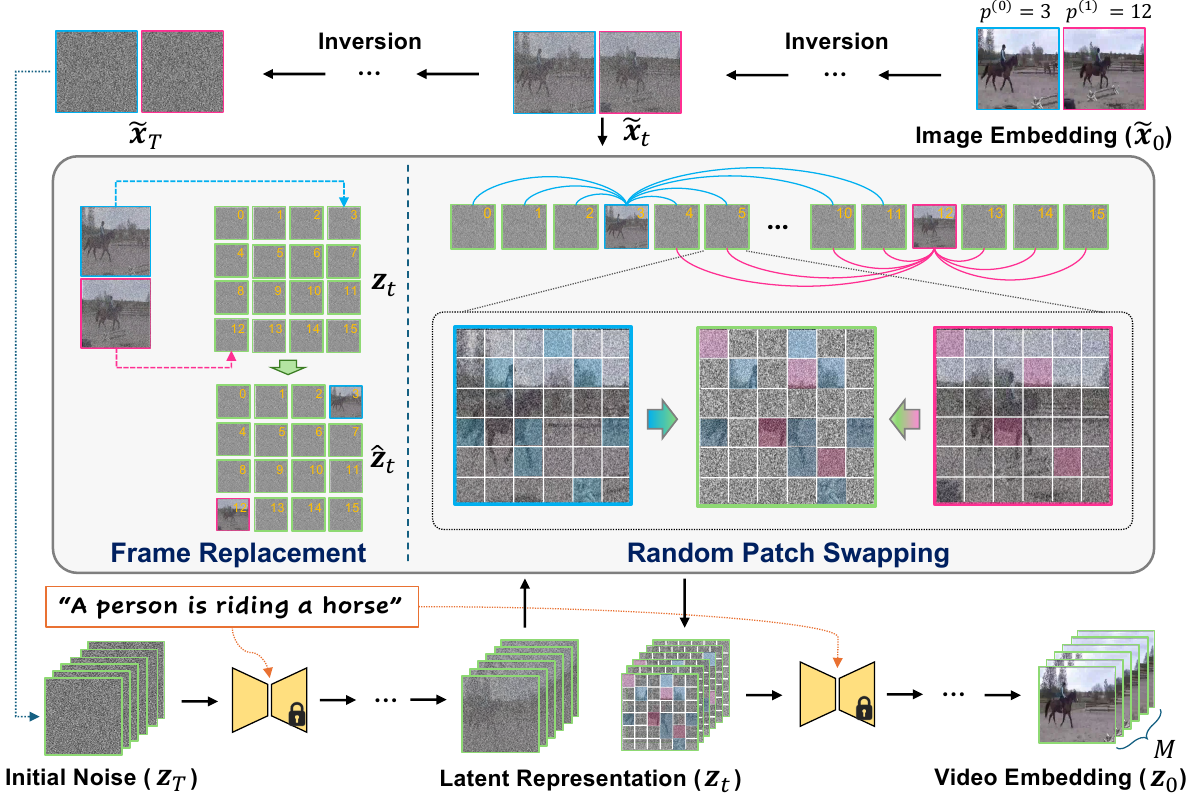}
  \caption{\textbf{Overview of the proposed FlexTI2V approach.} We invert the condition image embedding to noisy representation $\tilde{\bm{x}}_t$ at each step. The final noise $\tilde{\bm{x}}_T$ is reused as initialization for video synthesis. At step $t$, we directly replace the video frames with images at the desired positions. Then, for each video frame, we randomly swap a portion of patches with bounded condition images based on the relative distance between the frame and images. Though we show a special case of using two condition images in this figure, our method can naturally extend to any number of images at any positions. Note that all operations of our method occur in the latent space. We visualize RGB images on the latent representations simply for intuitive understanding.}
  \label{fig:method}
  \vspace{-0.1cm}
\end{figure}

\section{FlexTI2V}

The setting of the proposed generic text-image-to-video (TI2V) generation problem is illustrated in Fig. \ref{fig:i2v_tasks}. The input includes a textual prompt $y$, $N$ images $\bm{x}=\{x^{(n)}\}_{n=0}^{N-1}$ as visual conditions, and the expected position index $\bm{p}=\{p^{(n)}\}_{n=0}^{N-1}$ of each image in the output video. The goal is to synthesize a video consisting of $M$ frames $\bm{v}=\{v^{(m)}\}_{m=0}^{M-1}$ with each condition image at the expected position (\textit{i.e.}, $v^{\left(p^{(n)}\right)}=x^{(n)}$ where $p^{(n)}\in [0,M-1]$). The output video is supposed to show a reasonable transition across condition images following the textual prompt. Our problem setting unifies many existing TI2V tasks in one formulation, such as image animation ($N=1$, $\bm{p}=\{0\}$), video rewinding ($N=1$, $\bm{p}=\{M-1\}$) and video frame interpolation ($N=2$, $\bm{p}=\{0,M-1\}$).

The major challenge of training-free TI2V generation is harmonizing the distinct denoising directions controlled by texts and images solely relying on frozen model parameters. The experiments of MAE \cite{he2022masked} validate that global image information can be carried by and recovered from a portion of local patches. Inspired by this finding, the key insight of our \textit{FlexTI2V} method is contextualizing each synthetic video frame with visual conditioning, by randomly swapping a portion of patches between video frames and condition images in the latent space at early denoising steps. To balance the fidelity to condition images and creativity in motion generation, we use a dynamic control strategy for the percentage of swapped patches and denoising steps for each video frame, based on their distance to the condition images. This strategy also introduces position information of condition images into the denoising process, thus making the diffusion model capable of handling multiple condition images at any desired positions.

\subsection{Condition Image Inversion and Replacement}

\textbf{Image Inversion}\quad The overview of our method is illustrated in Fig. \ref{fig:method}. First of all, we encode condition images $\bm{x}$ into latent embedding $\tilde{\bm{x}}$ using the pretrained encoder $\mathcal{E}$, \textit{e.g.}, $\tilde{\bm{x}}=\mathcal{E}(\bm{x})=\{\tilde{x}^{(n)}\}_{n=0}^{N-1}$. To fuse the visual condition with T2V denoising process, we invert the clean latent image embedding into noisy representations at each denoising step. We denote the representation at step $t=0$ as $\tilde{\bm{x}}_0=\tilde{\bm{x}}$ for consistent notations in diffusion models. The widely-used DDIM inversion \cite{mokady2023null,titov2024guide} is not applicable to our problem because the video diffusion model can not estimate the noise accurately in a single image without temporal information. Instead, we use the straightforward DDPM-based inversion as an alternative following \cite{meng2022sdedit,ni2024ti2v}, which directly adds Gaussian noise to $\tilde{\bm{x}}_0$ like diffusion process, $\tilde{\bm{x}}_t\sim\mathcal{N}\left(\sqrt{\bar{\alpha}_t}\;\tilde{\bm{x}}_0,\ (1-\bar{\alpha}_t)\bm{I}\right)$. In denoising process, T2V diffusion models typically initialize the noise by drawing a random sample from Gaussian distribution \cite{ho2020denoising}. Recent work has validated reusing noise inverted from images preserves the global structure in image editing \cite{tumanyan2023plug,mokady2023null,liu2024towards}, and leads to consistent video frames in video generation \cite{ni2024ti2v}. Hence, we duplicate the inverted noise $\tilde{\bm{x}}_T$ at the final step, and stack them along the temporal dimension as the initial $\bm{z}_T$ for video denoising.

\noindent\textbf{Frame Replacement}\quad As shown in Fig. \ref{fig:method}, the representation of the $M$ video frames at step $t$ is denoted as $\bm{z}_t=\{z_t^{(m)}\}_{m=0}^{M-1}$. We leverage the pretrained temporal modeling layer in the T2V diffusion model to propagate visual features of condition images to all video frames. Inspired by \cite{ni2024ti2v}, we replace the video frame embeddings at the conditioning position $\bm{p}=\{p^{(n)}\}_{n=0}^{N-1}$ with inverted image embeddings, which is formulated as
\begin{equation}\scriptsize
  \text{\texttt{FrameReplace}}\left(\tilde{\bm{x}}_t,\bm{z}_t,\bm{p}\right)= \left\{z_t^{(0)}, z_t^{(1)}, \cdots, z_t^{\left(p^{(n)}-1\right)}, \tilde{x}_t^{\left(p^{(n)}\right)}, z_t^{\left(p^{(n)}+1\right)}, \cdots, z_t^{(M-2)}, z_t^{(M-1)}\right\},
\end{equation}
where $n\in [0, N-1]$. We refer to the resulting video frame embedding as $\hat{\bm{z}}_t$.

\subsection{Random Patch Swapping}

Due to the limited number of condition images compared with the long video frame sequence, the features of condition images may be neglected in the temporal layers. The generated video thus still deviates from the visual conditioning. To tackle this challenge, we introduce a novel \textit{random patch swapping} (RPS) method to explicitly incorporate visual features into each video frame by swapping a random portion of patches between each pair of video frame embedding $\hat{z}_t^{(m)}$ and condition image embedding $\tilde{x}_t^{(n)}$ (see Fig. \ref{fig:method} for visualization). Specifically, we pre-define a percentage, $P(m,n,t)$, of swapped patches between $m$-th video frame and $n$-th condition image at step $t$. Then we generate a random binary mask $\mathcal{M}_t^{(m,n)}$ composed of zeros and ones having the same shape as $\hat{z}_t^{(m)}$ and $\tilde{x}_t^{(n)}$, $\textit{i.e.}$, $\mathcal{M}_t^{(m,n)}$, $\hat{z}_t^{(m)}$, $\tilde{x}_t^{(n)}\in\mathbb{R}^{C\times H\times W}$. The number of ones in the mask account for $P(m,n,t)$ of the total elements, so we have $\sum_{c,i,j}\mathcal{M}_t^{(m,n)}[c,i,j] / (C\times H\times W)=P(m,n,t)$. Note that the values are identical across the channel dimension at each spatial location $(i,j)$, so we omit the channel $c$ in following paragraphs for simplicity.

Random patch swapping is essentially replacing the element of $m$-th video frame at position $(i,j)$ with the element of $n$-th condition image at the same position. Each element in the latent space corresponds to a patch in RGB space depending on the encoder compression ratio. Then RPS can be formulated as updating the values in $\hat{z}_t^{(m)}$ following the equation below:
\begin{equation}
    \hat{z}_t^{(m)}[i,j] = \mathcal{M}_t^{(m,n)}[i,j] \cdot \tilde{x}_t^{(n)}[i,j] + \left(1-\mathcal{M}_t^{(m,n)}[i,j]\right) \cdot \hat{z}_t^{(m)}[i,j].
\end{equation}
In this way, the visual cues from condition images are delivered to video frames through swapped patches, and then harmonized with video features by the pretrained diffusion foundation models. Each video frame only swaps patches with the two bounded condition images (the closest condition images before and after it). If all condition images are on the same side of a video frame, this frame only swaps patches with the single nearest condition image. We find that a large percentage of swapped patches may include overly strong spatial alignment prior, while a small percentage contains insufficient visual features for effective guidance. To make a trade-off, we dynamically reduce the percentage and steps of RPS when the video frame is far from the position of condition image. The percentage $P(m,n,t)$ is written as
\begin{equation}\small
        P(m,n,t) = \left\{
        \begin{array}{ll}
             P_0 - \delta_1\cdot|m-p^{(n)}|, & T - \tilde{t} < t \leq T, \\
             0, & 0 \leq t \leq T - \tilde{t},
        \end{array}
        \right. \ \mathrm{where}\quad \tilde{t} = t_0 - \delta_2\cdot |m-p^{(n)}|.
\end{equation}
$P_0$ and $t_0$ are initial swapping percentage and steps, and $\delta_1$, $\delta_2$ are decreasing strides. Our dynamic control strategy is based on the intuitive hypothesis that video frames around condition images are also perceptually similar, thus demanding more visual cues by patch swapping. Likewise, for frames far from condition images, we reduce constraints of visual conditioning accordingly to encourage more creativity in motion generation. We only swap patches in the first $\tilde{t}$ denoising steps, and let the model to harmonize and smooth artifacts in the remaining $T-\tilde{t}$ steps by setting $P(m,n,t)=0$. Additionally, This method also introduces position information of images into a T2V model, which makes our approach applicable to any number of images at any positions. The entire algorithm formulation is shown in the supplementary.

\section{Experiments}

\subsection{Dataset and Metrics}
\label{subsec:dataset_metrics_implement}

\textbf{Dataset} We run experiments on the widely-used UCF-101 dataset \cite{soomro2012ucf101} following \cite{ho2022video,fu2023tell,ni2024ti2v}, which is composed of 13,320 videos of 101 action categories. We select 42 action classes including complex actions with high dynamics (\textit{e.g.}, surfing) and small motion with still background (\textit{e.g.}, typing). For each class, we randomly select 25 instances (totaling 1,050 videos) as the test set. We manually rewrite the textual prompts for each action. We also run experiments on one more synthetic dataset to consolidate our conclusion, which is shown in the supplementary.

\noindent\textbf{Metrics}\quad We use three CLIP-based metrics including (1) CLIP frame-text similarity (CLIP-T) measuring the fidelity of generated frames to textual prompts, (2) CLIP frame similarity (CLIP-F) measuring the consistency across all video frames, and (3) CLIP image similarity (CLIP-I) measuring the agreement between condition images and video frames at other positions. We also use (4) Frechet Video Distance (FVD) \cite{unterthiner2018towards} to assess the overall video quality, its variant (5) text-conditioned FVD (tFVD) \cite{ni2024ti2v} to assess the video-text alignment, and (6) user study for human evaluation.

\noindent\textbf{Implementation Details}\quad For a fair comparison with \cite{ni2024ti2v}, we implement our FlexTI2V method on ModelScopeT2V \cite{wang2023modelscope}, which shows great robutness to out-domain T2V generation. We set the number of video frames as $M=16$, and total denoising steps $T=20$ with deterministic DDIM sampling (\textit{i.e.}, $\sigma_t=0$ in Eq. \ref{eq:ddim}). The initial percentage and steps of swapping are set as $P_0=0.3$ and $t_0=10$. The decreasing strides are $\delta_1=5\times10^{-3}$ and $\delta_2=0.3$. Our method can be run on 1 NVIDIA H100 GPU. Please refer to the supplementary for more implementation details.

\begin{table}[t]
\small
\centering
\setlength{\tabcolsep}{0.23cm}
\begin{tabular}{lcccccc}
\toprule
Methods & User & FVD$\downarrow$ & tFVD$\downarrow$ & CLIP-T & CLIP-F & CLIP-I \\
\midrule

\multicolumn{7}{c}{\textit{Image Animation}} \\
\midrule
DynamiCrafter \cite{xing2024dynamicrafter} & 26.20 & 210.77 & 832.38 & 29.86 & 89.10 & 80.95 \\
TI2V-Zero \cite{ni2024ti2v}                & 28.40 & 189.49 & 586.54 & 29.84 & 88.01 & 82.29 \\
\rowcolor[HTML]{FAEBD7} \textbf{FlexTI2V}  & \textbf{45.40} & \textbf{125.49} & \textbf{558.29} & \textbf{30.97} & \textbf{94.40} & \textbf{84.38} \\
\midrule

\multicolumn{7}{c}{\textit{Rewinding}} \\
\midrule
TI2V-Zero \cite{ni2024ti2v}                & 39.80 & 199.01 & 611.60 & 29.77 & 89.21 & 81.02 \\
\rowcolor[HTML]{FAEBD7} \textbf{FlexTI2V}  & \textbf{60.20} & \textbf{134.98} & \textbf{596.96} & \textbf{30.95} & \textbf{94.58} & \textbf{85.11} \\
\midrule

\multicolumn{7}{c}{\textit{Outpainting}} \\
\midrule
TI2V-Zero \cite{ni2024ti2v}                & 38.60 & 216.80 & 641.07 & 28.80 & 89.01 & 85.01 \\
\rowcolor[HTML]{FAEBD7} \textbf{FlexTI2V}  & \textbf{61.40} & \textbf{167.46} & \textbf{632.71} & \textbf{30.75} & \textbf{95.89} & \textbf{88.34} \\
\midrule

\multicolumn{7}{c}{\textit{Video Frame Interpolation}} \\
\midrule
DynamiCrafter \cite{xing2024dynamicrafter} & 16.20 & 310.74 & 731.90 & 29.69 & 84.26 & 78.68 \\
TRF \cite{feng2024explorative}             & 32.20 & 247.88 & 631.54 & 29.47 & 91.99 & 83.35 \\
\rowcolor[HTML]{FAEBD7} \textbf{FlexTI2V}  & \textbf{51.60} & \textbf{98.41} & \textbf{406.30} & \textbf{30.92} & \textbf{92.34} & \textbf{86.88} \\
\bottomrule
\end{tabular}
\caption{\textbf{Quantitative comparison with previous methods.} \colorbox{lightgray}{``User''} denotes the user study results. $\downarrow$ means a lower score indicates better performance. The \textcolor{orange}{orange} row refers to our method.}
\vspace{-0.4cm}
\label{tab:sota_cmp}
\end{table}

\subsection{Quantitative Comparison}
\label{sec:cmp_sota}

We compare with prior approaches quantitatively under three classic settings: (1) image animation -- one condition image as the first video frame ($N=0$, $\bm{p}=\{0\}$), (2) rewinding -- one condition image as the last video frame ($N=0$, $\bm{p}=\{15\}$), and (3) video frame interpolation -- two condition images as the first and last frames ($N=2$, $\bm{p}=\{0,15\}$). We also introduce a new setting, video frame outpainting -- one condition image as the middle video frame ($N=1$, $\bm{p}=\{7\}$). We mainly compare with two state-of-the-art \textit{training-free} methods -- TI2V-Zero \cite{ni2024ti2v} and time reversal fusion (TRF) \cite{feng2024explorative}. TI2V-Zero is initially designed only for image animation task. We adapt it to outpainting and rewinding tasks by adjusting the order of frame synthesis. Moreover, we also include DynamiCrafter \cite{xing2024dynamicrafter} as a \textit{training-based} baseline for a thorough comparison. More details of baseline implementation are elaborated in the supplementary. The results are shown in Tab. \ref{tab:sota_cmp}.

\noindent\textbf{Single-Image Conditioning}\quad For image animation, outpainting and rewinding tasks, our method outperforms DynamiCrafter and TI2V-Zero by a notable margin in FVD, which suggests an overall higher quality of videos synthesized by our method. In fine-grained evaluation metrics, our FlexTI2V method surpasses the best baseline by 28.25/8.36/14.64 in tFVD and 1.11\%/1.95\%/1.18\% in CLIP-T for the three tasks, indicating the best prompt-following capability. In addition, the two baselines have low scores in CLIP-F and CLIP-I, indicating the inconsistency across generated video frames and low fidelity to the condition images. The reason could be that DynamiCrafter is a training-based model which still suffers from the bottleneck of generalization to out-of-domain data. Though TI2V-Zero is a training-free method, it synthesizes the future video frames autoregressively, resulting in compounding errors and cumulative deviations from visual conditions, which can also be observed in visualization in Sec. \ref{sec:visualization}. In contrast, our method generates all video frames simultaneously in a single denoising process, thus achieving the best performance in CLIP-F and CLIP-I. The prominent improvement (5.30\%/6.88\%/5.37\% and 2.09\%/3.33\%/4.09\%) also validates the effectiveness of our random patch swapping strategy for various image conditioning settings.

\noindent\textbf{Multi-Image Conditioning}\quad In terms of video frame interpolation, DynamiCrafter still lags behind TRF and FlexTI2V due to domain shift. In contrast, TRF shows better performance for this task (measured by FVD) relying on the strong generalization capability of stable video diffusion \cite{blattmann2023stable}, which further validates the advantage of training-tree method in controllable video generation. Though TRF achieves a high score in CLIP-F and CLIP-I, it still has a limited performance in video-text alignment (tFVD and CLIP-T) due to lack of semantic control by textual prompts. In contrast, our method outperforms both competitors in all the metrics. The significant improvement further validates the flexibility of our method to multi-image conditioning.

\begin{table}[t]
\footnotesize
\centering
\setlength{\tabcolsep}{0.15cm}
\begin{tabular}{lcccc}
\toprule
& DynamiCrafter \cite{xing2024dynamicrafter} & TI2V-Zero \cite{ni2024ti2v} & TRF \cite{feng2024explorative} & \cellcolor[HTML]{FAEBD7} \textbf{FlexTI2V} \\
\midrule
Inference Time (s)   & 9.45 & 275.14 & 60.05 & \cellcolor[HTML]{FAEBD7} \textbf{2.71} \\
\bottomrule
\end{tabular}
\caption{\textbf{Comparison of model efficiency in inference.} The \textcolor{orange}{orange} column refers to our method. Our method shows great advantage in inference speed compared with other baseline models.}
\label{tab:speed}
\vspace{-0.4cm}
\end{table}

\noindent\textbf{User Study}\quad Besides automatic metrics. We also conduct user study for a thorough evaluation. We sample 100 instances from our test set for each task, with each instance evaluated by five raters. Each rater is asked to select the video with best alignment with the textual prompt and condition images. The human preference percentage of each method is presented in Tab. \ref{tab:sota_cmp}. FlexTI2V is selected as the best in the highest percentage of samples compared with other competitors. The results suggest that the videos synthesized by our method align better with human's subjective perception.

\noindent\textbf{Inference Efficiency}\quad We evaluate the average inference time of our method and the baseline models for generating a single 16-frame video in Tab. \ref{tab:speed}. The test is conducted on one NVIDIA H100 GPU. TI2V-Zero synthesizes video frames one by one, and relies on multiple resample steps to improve the performance. The inference time thus is significantly longer than other approaches. TRF conducts denoising processes conditioned on two bounded images separately and then merges the features at each step, resulting in longer inference time than DynamiCrafter. In contrast, our method requires only 2.71 seconds for 20 DDIM denosing steps, which is much lower than DynamiCrafter and TRF. The great gap validates the high efficiency of our method.

\begin{figure}[t]
  \centering
  \includegraphics[width=\linewidth]{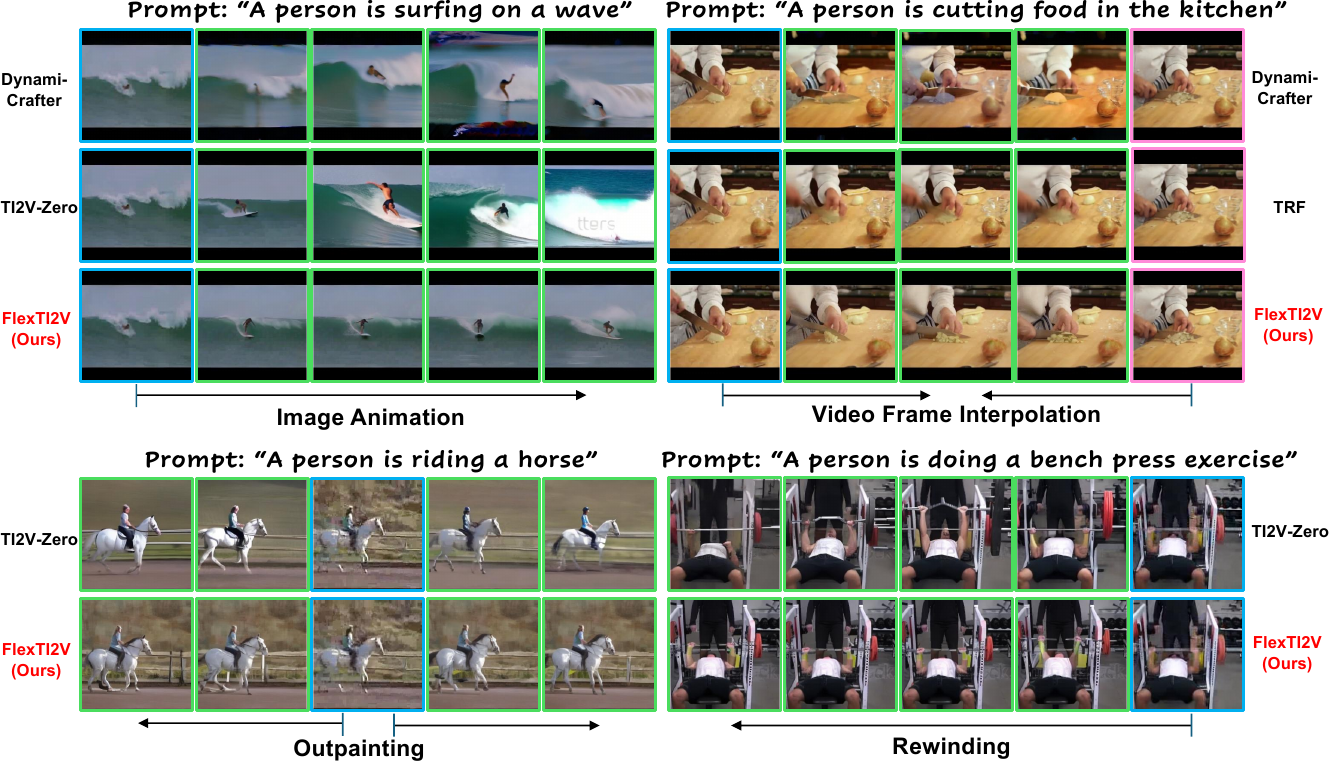}
  \caption{\textbf{Comparison with prior methods.} The images with \textcolor{NavyBlue}{blue} and \textcolor{Lavender}{pink} edges are condition images for each setting, and images with \textcolor{ForestGreen}{green} edges are generated video frames. Our approach synthesizes videos with higher frame consistency and fidelity to condition images than other baseline models in various settings.}
  \label{fig:cmp_sota}
\end{figure}

\subsection{Visualization for Qualitative Comparison}
\label{sec:visualization}

Besides quantitative evaluation, we also visualize the videos synthesized by our method and other baselines for a qualitative comparison. As illustrated in Fig. \ref{fig:cmp_sota}, we observe some distortion in the videos generated by DynamiCrafter (\textit{e.g.}, the person in image animation and the knife in video frame interpolation), which may be the explanation for the low performance in Tab. \ref{tab:sota_cmp}. TRF is good at preserving visual features of the two bounded images. However, the video frames synthesized by TRF are blurry around the knife. We speculate that TRF is initially proposed for smooth motion generation and camera viewpoint transition. It may be weak in modeling object state change, such as the onion being cut into small pieces in this example. TI2V-Zero can generate videos with large motion because of the autoregressive generation mechanism. However, the cumulative mistakes weaken the controllability of visual conditions, leading to a low consistency in the frames far from the condition image. In contrast, our method can follow the textual prompts and visual conditions faithfully, and synthesize smooth motions in all TI2V settings. More examples of our method are illustrated in the supplementary.
\vspace{-0.1cm}

\begin{table}[t]
\small
\centering
\setlength{\tabcolsep}{0.45cm}
\begin{tabular}{lccc}
\toprule
Methods & FVD$\downarrow$ & CLIP-T & CLIP-I \\
\midrule
\multicolumn{4}{c}{\textit{Image Animation}} \\
\midrule
Wan2.1-I2V \cite{wang2025wan} & 141.22 & 31.01 & 82.33 \\
\rowcolor[HTML]{FAEBD7} Wan2.1-T2V + Our Method & \textbf{101.74} & \textbf{30.98} & \textbf{85.87} \\
\midrule
\multicolumn{4}{c}{\textit{Video Frame Interpolation}} \\
\midrule
Wan2.1-FLF2V \cite{github2025wan} & 112.22 & 31.47 & 85.13 \\
\rowcolor[HTML]{FAEBD7} Wan2.1-T2V + Our Method & \textbf{90.82} & \textbf{31.50} & \textbf{86.04} \\
\bottomrule
\end{tabular}
\caption{\textbf{Comparison with Wan2.1 model family.} $\downarrow$ means a lower score in this metric indicates a better performance. The \textcolor{orange}{orange} column refers to our method.}
\label{tab:cmp_wan}
\vspace{-0.5cm}
\end{table}

\begin{figure}[t]
  \centering
  \includegraphics[width=\linewidth]{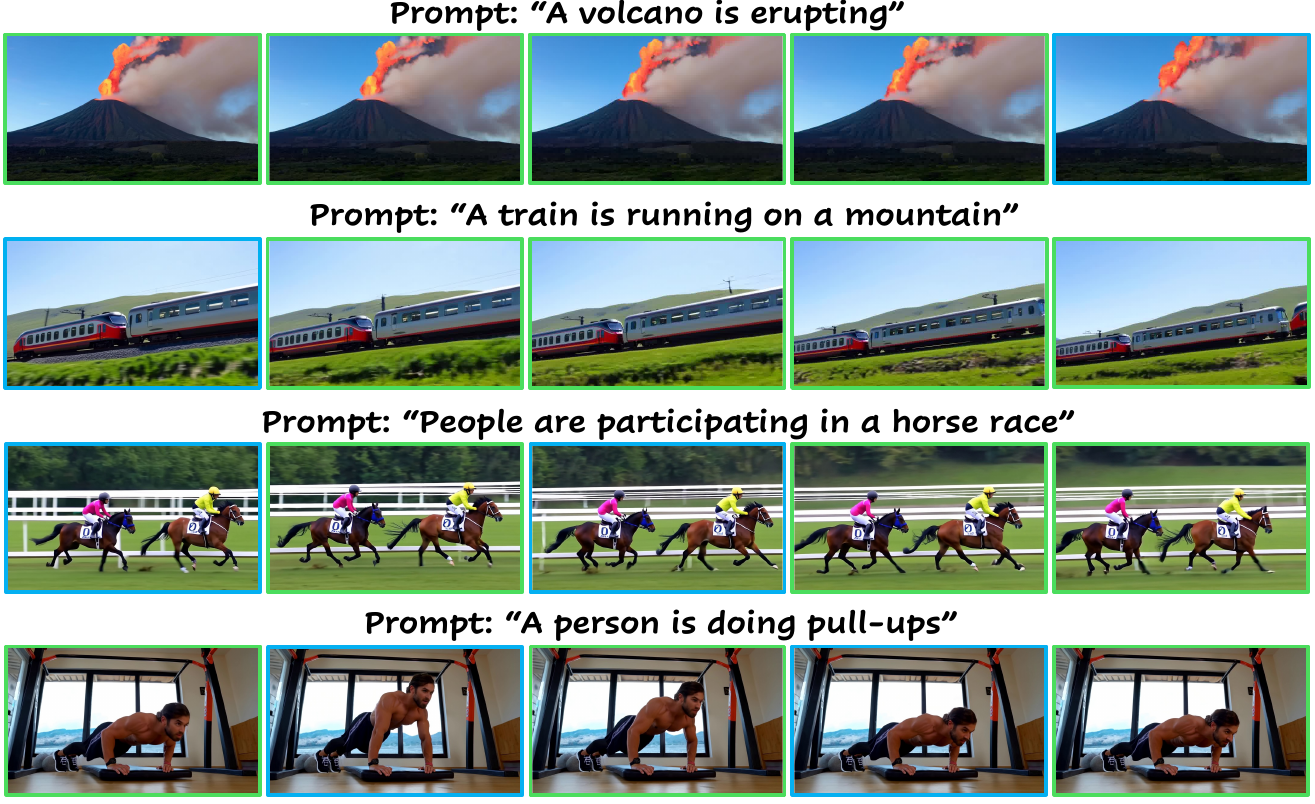}
  \caption{\textbf{Generated videos of our method implemented to Wan2.1-T2V.} The images with \textcolor{NavyBlue}{blue} edges are condition images and images with \textcolor{ForestGreen}{green} edges are generated video frames. Please refer to supplementary for corresponding videos and extra demos.}
  \label{fig:wan}
  \vspace{-0.2cm}
\end{figure}

\subsection{Generalization to Different Foundation Models}

We use ModelScopeT2V \cite{wang2023modelscope} as the foundation model in previous experiments simply for a fair comparison with prior methods. To validate the generalization of our method, we also implement our method to another SOTA T2V foundation model -- Wan2.1-T2V \cite{wang2025wan} using transformer-based architectures. We directly compare with Wan2.1-I2V and Wan2.1-FLF2V on image animation and video frame interpolation separately. The two specialized models are finetuned from the base Wan2.1-T2V model for the two specific tasks, which makes it an apple-to-apple comparison with our proposed approach.

As shown in \cref{tab:cmp_wan}, our method outperforms the two specialized models on the two tasks. The possible reason is that it’s still challenging for training-based methods to generalize to out-of-distribution data (\eg, different aspect ratios and resolutions). This also shows the necessity of training-free methods like our work. We demonstrate videos synthesized by our method in \cref{fig:wan}. Implemented on top of the modern architecture, our method generates videos with high fidelity and smooth motions, showing robust generalization to different T2V models. 

\begin{table}[t]
\small
\centering
\setlength{\tabcolsep}{0.08cm}
\begin{tabular}{lcccccccccc}
\toprule
\multirow{2.5}{*}{}  & \multicolumn{2}{c}{Animation} & \multicolumn{2}{c}{Outpainting} & \multicolumn{2}{c}{Rewinding} & \multicolumn{2}{c}{Interpolation} \\
\cmidrule(lr){2-3} \cmidrule(lr){4-5}  \cmidrule(lr){6-7} \cmidrule(lr){8-9}
& FVD$\downarrow$ & CLIP-I & FVD$\downarrow$ & CLIP-I & FVD$\downarrow$ & CLIP-I & FVD$\downarrow$ & CLIP-I  \\
\midrule
\rowcolor[HTML]{FAEBD7} FlexTI2V        & \textbf{125.49} & \textbf{84.38} & \textbf{167.46} & \textbf{88.34} & \textbf{134.98} & \textbf{85.11} & \textbf{98.41} & \textbf{86.88} \\
w/o Frame Replace                       & 130.50 & 83.90 & 174.02 & 87.80 & 137.20 & 84.50 & 104.11 & 86.51 \\
w/o RPS                                 & 440.28 & 64.41 & 510.69 & 70.02 & 430.21 & 66.20 & 425.86 & 68.23 \\
\bottomrule
\end{tabular}
\caption{\textbf{Ablation study of components in our method.} $\downarrow$ means a lower score in this metric indicates a better performance. The \textcolor{orange}{orange} row refers to our complete method. Each component is removed from the complete method \textit{separately}.}
\vspace{-0.4cm}
\label{tab:ablation}
\end{table}

\subsection{Ablation Study}

In our method, there are two components (frame replacement and random patch swapping) to incorporate visual conditioning into T2V diffusion models. We evaluate the contribution of each component by ablating them from our method \textit{separately}. The results are presented in Tab. \ref{tab:ablation}. After removing frame replacement, the performance lags behind the full model by a small margin in all tasks. The result indicates that random patch swapping (RPS) can work alone to effectively merge visual guidance into T2V diffusion process. It also validates our hypothesis that sufficient visual features can be transferred through a portion of local patches at each denoising step for video generation. On the contrary, if we remove RPS and only implement frame replacement strategy for extra image conditioning, a significant performance drop is observed across all metrics and tasks. Though frame replacement introduces image features into the temporal layers, the model estimates the holistic noise added to all video frames (16 frames in our experimental setting) as an entirety. Hence, the condition image features are likely to be ignored because of the small number of images against the large number of video frames. We observe that the videos generated without RPS greatly deviate from the condition images, thus resulting in a remarkable decrease in the performance. We also visualize examples of the ablation study for qualitative analysis. Please check out our supplementary.

\begin{table}[t]
\small
\centering
\begin{tabular}{cc}

\setlength{\tabcolsep}{0.1cm}
\begin{tabular}{cccc}
\toprule
$P_0$ & $0.2$ & \cellcolor[HTML]{FAEBD7} $\bm{0.3}$ & $0.4$ \\
\midrule
FVD$\downarrow$ & 156.64 & \cellcolor[HTML]{FAEBD7} \textbf{125.49} & 147.11 \\
\bottomrule
\end{tabular}
&
\setlength{\tabcolsep}{0.05cm}
\begin{tabular}{ccccc}
\toprule
$\delta_1$ & $4\times10^{-3}$ & \cellcolor[HTML]{FAEBD7} $\bm{5\times10^{-3}}$ & $6\times10^{-3}$ & $7\times10^{-3}$ \\
\midrule
FVD$\downarrow$ & 133.79 & \cellcolor[HTML]{FAEBD7} \textbf{125.49} & 137.46 & 151.70 \\
\bottomrule
\end{tabular}
\\[0.5cm]

\setlength{\tabcolsep}{0.11cm}
\begin{tabular}{cccc}
\toprule
$t_0$ & $7$ & \cellcolor[HTML]{FAEBD7} $\bm{10}$ & $13$ \\
\midrule
FVD$\downarrow$ & 136.65 & \cellcolor[HTML]{FAEBD7} \textbf{125.49} & 134.96 \\
\bottomrule
\end{tabular}
&
\setlength{\tabcolsep}{0.21cm}
\begin{tabular}{ccccc}
\toprule
$\delta_2$ & $0.2$ & \cellcolor[HTML]{FAEBD7}$\bm{0.3}$ & $0.4$ & $0.5$ \\
\midrule
FVD$\downarrow$ & 133.79 & \cellcolor[HTML]{FAEBD7} \textbf{125.49} & 137.46 & 151.70 \\
\bottomrule
\end{tabular}
\\

\end{tabular}
\caption{\textbf{Hyperparameter analysis.} We change only one hyperparameter in each subtable and keep the other hyperparameters fixed. $\downarrow$ means a lower score indicates better performance. The \textcolor{orange}{orange} columns refer to the hyperparameters we finally adopt.}
\label{tab:hyperparameter_ablation}
\vspace{-0.3cm}
\end{table}

\subsection{Hyperparameter Analysis}

In the proposed RPS approach, there are four important hyperparameters -- initial swapping percentage $P_0$, initial number of swapping steps $t_0$ and the decreasing strides $\delta_1$, $\delta_2$ for swapping percentage and steps. The optimal hyperparameters are selected empirically based on the analysis in \cref{tab:hyperparameter_ablation}. If we use a large $\delta_1$ and $\delta_2$, or small $P_0$ and $t_0$, the image conditioning is weaker than the optimal hyperparameters so that the generated video can not accurately follow the images. Likewise, if $\delta_1$ and $\delta_2$ are small, or $P_0$ and $t_0$ are large, the image conditioning is too strong so that the output videos are nearly frozen. To achieve an optimal balance, we select the best hyperparameter combination based on the experiments. We also find the optimal combination is very consistent across different datasets, so we use it in all the experiments.

\section{Conclusion}

In this paper, we introduce a generic text-image-to-video (TI2V) problem that unifies existing TI2V tasks, which requires the model be able to synthesize smooth videos conditioned on an arbitrary number of images at any expected positions. To solve this problem, we propose a novel training-free method, FlexTI2V, that can introduce flexible image conditioning to pretrained text-to-video diffusion models. The key component of our method is random patch swapping strategy, which merges visual features of condition images into video denoising process by swapping a random portion of local patches. The extensive experiments indicate a notable improvement of FlexTI2V over prior methods, as well as the flexibility of our model. Our work is an important step for flexible and low-cost controllable video generation, which provides an alternative way to adapt pretrained foundation models to specific downstream tasks, thus democratizing the access to customized video generation tools.

%
%
\bibliographystyle{splncs04}
\bibliography{main}


\clearpage
\appendix

\begin{center}
    \textbf{\Large Unified Text-Image-to-Video Generation: \\ A Training-Free Approach to Flexible Visual Conditioning}
    \\ [0.8cm]
    {\Large Supplementary Material}
    \\ [1.2cm]
\end{center}

This is the supplementary material for the paper titled ``Unified Text-Image-to-Video Generation: A Training-Free Approach to Flexible Visual Conditioning''. We organize the content as follows:
\\

\noindent\textbf{\hyperref[sec:algorithm]{A} -- Algorithm Formulation} \\ [0.2cm]
\textbf{\hyperref[sec:additional_experiments]{B} -- Additional Experiment Results} \\ [0.1cm]
\indent\hyperref[sec:synthetic_dataset]{B.1} -- Quantitative Results on Synthetic Dataset  \\ [0.1cm]
\indent\hyperref[sec:dynamic_control]{B.2} -- Analysis of Dynamic Control Strategy \\ [0.1cm]
\indent\hyperref[sec:ablation_demo]{B.3} --  Visualization for Ablation Study \\ [0.1cm]
\indent\hyperref[sec:additional_nonhuman]{B.4} -- Generated Videos without Humans  \\ [0.1cm]
\indent\hyperref[sec:additional_demo]{B.5} -- Additional Demonstration of Our Method  \\ [0.1cm]
\indent\hyperref[sec:failures]{B.6} -- Failure Cases  \\ [0.2cm]
\textbf{\hyperref[sec:more_implementation]{C} -- Implementation Details} \\ [0.1cm]
\indent\hyperref[sec:test_set]{C.1} -- Establishment of Test Set \\ [0.1cm]
\indent\hyperref[sec:our_method_details]{C.2} -- More Details of Our Method \\ [0.1cm]
\indent\hyperref[sec:implementation_previous_methods]{C.3} -- Implementation of Previous Methods \\ [0.2cm]
\textbf{\hyperref[sec:limitation_future_work]{D} -- Limitation and Future Work} \\

\setcounter{section}{0}
\setcounter{subsection}{0}
\setcounter{figure}{0}
\setcounter{table}{0}
\setcounter{algorithm}{0}

\renewcommand\thesection{\Alph{section}}
\renewcommand\thesubsection{\thesection.\arabic{subsection}}
\renewcommand\thefigure{\Alph{figure}}
\renewcommand\thetable{\Alph{table}}
\renewcommand\thealgorithm{\Alph{algorithm}}

\renewcommand{\theHsection}{appendix.\Alph{section}}
\renewcommand{\theHsubsection}{appendix.\Alph{section}.\arabic{subsection}}
\renewcommand{\theHfigure}{appendix.\Alph{figure}}
\renewcommand{\theHtable}{appendix.\Alph{table}}
\renewcommand{\theHalgorithm}{appendix.\Alph{algorithm}}

\begin{algorithm}[t]
\scriptsize
\setstretch{1.2}
\caption{FlexTI2V}
\label{alg:flexti2v}
\begin{algorithmic}[1]
\Require The condition images $\bm{x}$; The expected positions of each image in the output video $\bm{p}$; The textual prompt $y$; A pretrained text-to-video diffusion foundation model, including a paired encoder $\mathcal{E}$ and decoder $\mathcal{D}$ as well as noise estimation network $\bm{\varepsilon}_\theta$. Hyperparameters including the initial percentage $P_0$, number of denoising steps $t_0$, the decreasing strides $\delta_1$ and $\delta_2$, and the amount of video frames $M$.
\Ensure A generated video of $M$ frames $\bm{v}$, with each condition image appearing at the expected position, \textit{i.e.}, $v^{(p^{(n)})}=x^{(n)},\ \forall n\in[0,N-1]$.
\State $\tilde{\bm{x}}_0 \gets \mathcal{E}(\bm{x})$
\State $\tilde{\bm{x}}_T\sim\mathcal{N}\left(\sqrt{\bar{\alpha}_T}\;\tilde{\bm{x}}_0,\ (1-\bar{\alpha}_t)\bm{I}\right)$
\State $\bm{z}_T \gets \verb+Repeat+(\tilde{x}_T^{(0)},\ M)$ \Comment{\textit{\textcolor{gray}{Noise Initialization}}}
\For{$t\in\{T, T-1, \cdots, 2, 1\}$}
    \State $\tilde{\bm{x}}_t\sim\mathcal{N}\left(\sqrt{\bar{\alpha}_t}\;\tilde{\bm{x}}_0,\ (1-\bar{\alpha}_t)\bm{I}\right)$ \Comment{\textit{\textcolor{gray}{Image Inversion}}}
    \State $\hat{\bm{z}}_t \gets \verb+FrameReplace+(\tilde{\bm{x}}_t,\ \bm{z}_t,\ \bm{p})$ \Comment{\textit{\textcolor{gray}{Frame Replacement, Eq. 3 in Main Paper}}}
    \For{$m\in\{0,1\cdots,M-1\}$}
        \State $\bm{n}_{bound} \gets \verb+BoundIndex+(m,\bm{p})$ \Comment{\textit{\textcolor{gray}{Indices of Bounded Condition Image}}}
        \For{$n\in\bm{n}_{bound}$}
            \State $\tilde{t} \gets t_0-\delta_2\cdot|m-p^{(n)}|$ \Comment{\textit{\textcolor{gray}{Dynamic Control in Swapping Steps}}}
            \If{$T-\tilde{t}<t\leq T$}
                \State $P(m,n,t) \gets P_0 - \delta_1\cdot|m-p^{(n)}|$ \Comment{\textit{\textcolor{gray}{Dynamic Control in Swapping Percentage}}}
            \Else
                \State $P(m,n,t) \gets 0$
            \EndIf
            \State Generate random binary mask $\mathcal{M}_t^{(m,n)}$ with element ``$1$'' accounting for $P(m,n,t)$.
            \For{$i\in\{0,1,\cdots,H-1\}, j\in\{0,1,\cdots,W-1\}$} \Comment{\textit{\textcolor{gray}{Random Patch Swapping}}}
                \State $\hat{z}_t^{(m)}[i,j] \gets \mathcal{M}_t^{(m,n)}[i,j] \cdot \tilde{x}_t^{(n)}[i,j] + \left(1-\mathcal{M}_t^{(m,n)}[i,j]\right) \cdot \hat{z}_t^{(m)}[i,j]$
            \EndFor
        \EndFor
    \EndFor
    \State $\bm{z}_t \gets \hat{\bm{z}}_t$
    \State $\bm{z}_{t-1} \gets \frac{\sqrt{\bar{\alpha}_{t-1}}}{\sqrt{\bar{\alpha}_t}} \left(\bm{z}_t-\sqrt{1-\bar{\alpha}_t}\cdot\bm{\varepsilon}_\theta(\bm{z}_t,t)\right) + \sqrt{1-\bar{\alpha}_{t-1}}\cdot\bm{\varepsilon}_\theta(\bm{z}_t,t)$ \Comment{\textit{\textcolor{gray}{Deterministic Sampling}}}
\EndFor
\State $\bm{v} \gets \mathcal{D}(\bm{z}_0)$
\State \Return $\bm{v}$
\end{algorithmic}
\end{algorithm}

\section{Algorithm Formulation}
\label{sec:algorithm}

We formulate our FlexTI2V approach in Alg. \ref{alg:flexti2v}. The function $\texttt{Repeat}(\tilde{x}_T^{(0)},M)$ at step 3 duplicates the inverted condition image representation for $M$ times as the initialization for video frame denoising. If there are more than one condition images, we find using different image makes trivial difference in the generation performance. We use the first condition image in our method. The function $\verb+BoundIndex+(m,\bm{p})$ at step 8 returns the indices of the \textit{two} closest condition images before and after $m$-th video frame respectively (\textit{i.e.}, bounded index). If all condition images are before or after this video frame, the function returns only \textit{one} index of the closest image.

\begin{table}[t]
\centering
\setlength{\tabcolsep}{0.4cm}
\begin{tabular}{lcccc}
\toprule
\multirow{2.5}{*}{Methods} & \multicolumn{2}{c}{Animation} & \multicolumn{2}{c}{Interpolation} \\
\cmidrule(lr){2-3} \cmidrule(lr){4-5}
& CLIP-T & CLIP-I & CLIP-T & CLIP-I \\
\midrule
DynamiCrafter \cite{xing2024dynamicrafter} & 24.33 & 79.53 & 25.44 & 75.50 \\
TRF \cite{feng2024explorative} & - & - & 25.92 & 81.38 \\
TI2V-Zero \cite{ni2024ti2v} & 24.87 & 81.76 & - & - \\
\rowcolor[HTML]{FAEBD7} FlexTI2V & \textbf{26.83} & \textbf{82.31} & \textbf{27.29} & \textbf{83.54} \\
\bottomrule
\end{tabular}
\caption{\textbf{Results on the synthetic dataset.} The \textcolor{orange}{orange} column refers to our method. Refer to \cref{sec:synthetic_dataset} for more descriptions.}
\label{tab:synthetic_dataset}
\end{table}

\section{Additional Experiment Resutls}
\label{sec:additional_experiments}

\subsection{Quantitative Results on Synthetic Dataset}
\label{sec:synthetic_dataset}

In the main paper, we follow the previous work \cite{ni2024ti2v} to conduct evaluation on UCF-101 for a fair comparison. To make the testing data more diverse, we utilize the data curation pipeline of \cite{ni2024ti2v} to synthesize an OPEN dataset by Stable Video Diffusion using 10 prompts:
\begin{itemize}\footnotesize
    \item[-] \textit{``A mesmerizing display of the northern lights in the Arctic.''}
    \item[-] \textit{``A bustling street market in Marrakech with colorful textiles and spices.''}
    \item[-] \textit{``A futuristic cityscape with holographic advertisements and flying cars.''}
    \item[-] \textit{``A romantic gondola ride through the canals of Venice at sunset.''}
    \item[-] \textit{``A group of friends on a road trip, singing along to their favorite songs.''}
    \item[-] \textit{``A serene mountain cabin covered in a fresh blanket of snow.''}
    \item[-] \textit{``A thrilling skateboarder performing tricks in a skate park.''}
    \item[-] \textit{``A bustling night market in Bangkok with street food vendors and live music.''}
    \item[-] \textit{``A high-speed bullet train racing through a scenic countryside.''}
    \item[-] \textit{``A group of explorers uncovering the mysteries of an ancient temple in the jungle.''} 
\end{itemize}
Different from UCF-101, the 10 prompts are mainly about non-human motions and object state changes. The synthetic data also guarantees the base T2V model is not trained on it. We generate 22 videos for each prompt, totaling 220 testing samples.

As shown in \cref{tab:synthetic_dataset}, our method still prominently outperforms all the baseline approaches in CLIP-T (text-video alignment) and CLIP-I (image-video similarity). The results validate that our method is also superior in modeling different types of motions and videos in the TI2V task.

\begin{figure}[t]
\centering
\includegraphics[width=\linewidth]{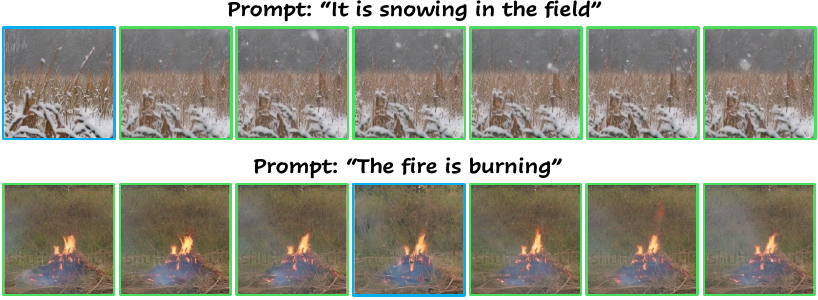}
\caption{Additional examples of our method on videos without humans. The images with \textcolor{NavyBlue}{blue} edges are condition images and images with \textcolor{ForestGreen}{green} edges are generated video frames. Please see Sec. \ref{sec:additional_nonhuman} for more details.}
\label{fig:additional_nonhuman}
\end{figure}

\begin{figure}[t]
\centering
\includegraphics[width=\linewidth]{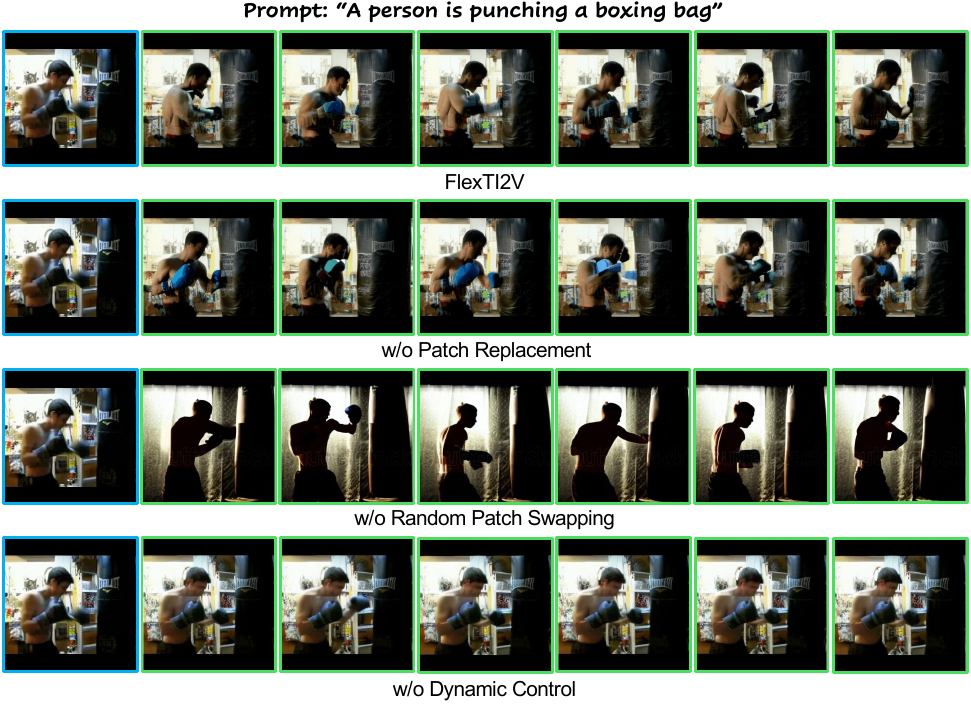}
\caption{Demonstration of ablation study. The images with \textcolor{NavyBlue}{blue} edges are condition images and images with \textcolor{ForestGreen}{green} edges are generated video frames. More explanations are provided in Sec. \ref{sec:dynamic_control} and Sec. \ref{sec:ablation_demo}.}
\label{fig:ablation}
\end{figure}

\subsection{Analysis of Dynamic Control Strategy}
\label{sec:dynamic_control}

We remove the dynamic control strategy from our method and use a fixed percentage and steps for random patch swapping. The results are illustrated in the last row of Fig. \ref{fig:ablation}. The motion in the generated video frames is trivial without dynamic control. The reason is because the constraints of visual conditioning are overly strong for all frames. The result validates the necessity of the dynamic control strategy.

\subsection{Visualization for Ablation Study}
\label{sec:ablation_demo}

Besides ablation of dynamic control (see Sec. \ref{sec:dynamic_control}), we also show additional visualization for quantitative ablation study of patch replacement and random patch swapping in Fig. \ref{fig:ablation}. After removing patch replacement, the output video is still decent with minor deviation from the condition image. However, without random patch swapping, the generated video fails to follow the visual condition, thus leading to a poor performance in main paper Tab. 4.

\begin{figure}[t]
  \centering
  \includegraphics[width=\linewidth]{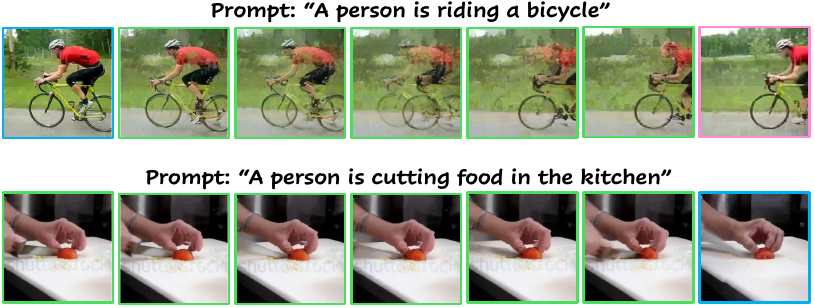}
  \caption{Typical failure cases of our approach. The images with \textcolor{NavyBlue}{blue} and \textcolor{Lavender}{pink} edges are condition images and images with \textcolor{ForestGreen}{green} edges are generated video frames. Please refer to Sec. \ref{sec:failures} for more analysis.}
  \label{fig:failures}
\end{figure}

\subsection{Generated Videos without Humans}
\label{sec:additional_nonhuman}

The videos in UCF-101 are mainly about human actions. To assess the generalization performance of our approach, we supplement extra examples of FlexTI2V synthesizing videos without humans in Fig. \ref{fig:additional_nonhuman}. Our approach seamlessly generalizes to the motion of objects showing strong generalization capability.

\begin{figure}[t]
  \centering
  \includegraphics[width=\linewidth]{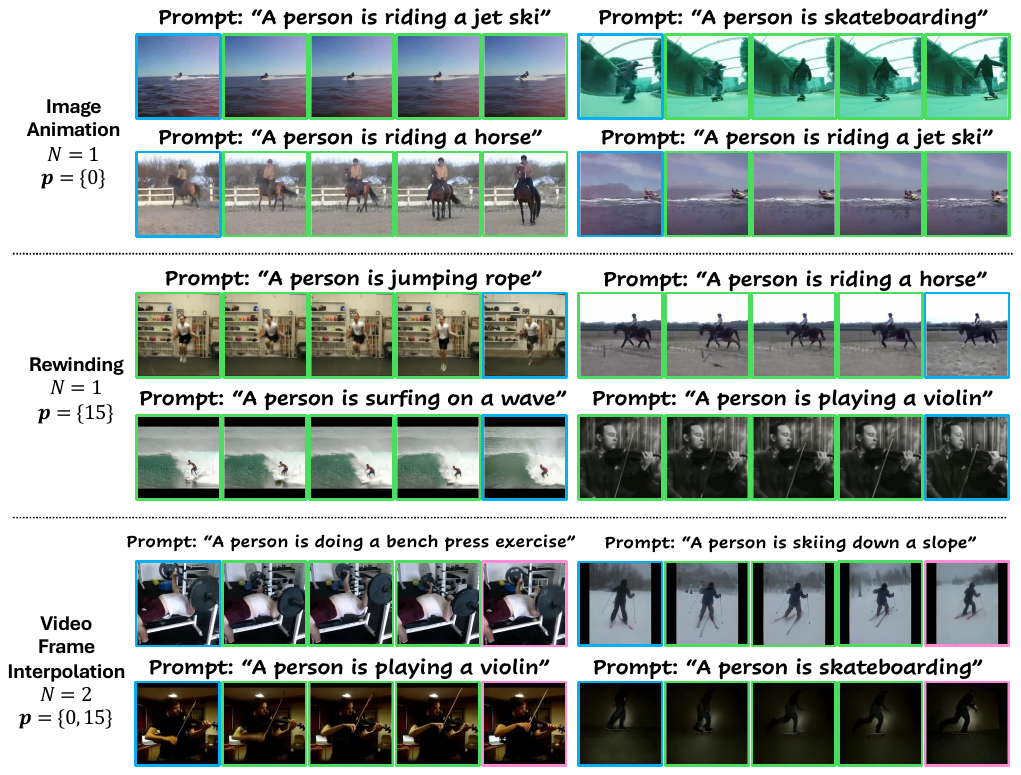}
  \caption{Additional demonstration of videos synthesized by FlexTI2V in classic TI2V settings. The images with \textcolor{NavyBlue}{blue} and \textcolor{Lavender}{pink} edges are condition imags, and images with \textcolor{ForestGreen}{green} edges are generated video frames. $N$ and $\bm{p}$ are the number and positions of condition images. Our method generates videos with 16 frames in the experiments.}
  \label{fig:additional_demo_classic}
\end{figure}

\begin{figure}[t]
  \centering
  \includegraphics[width=\linewidth]{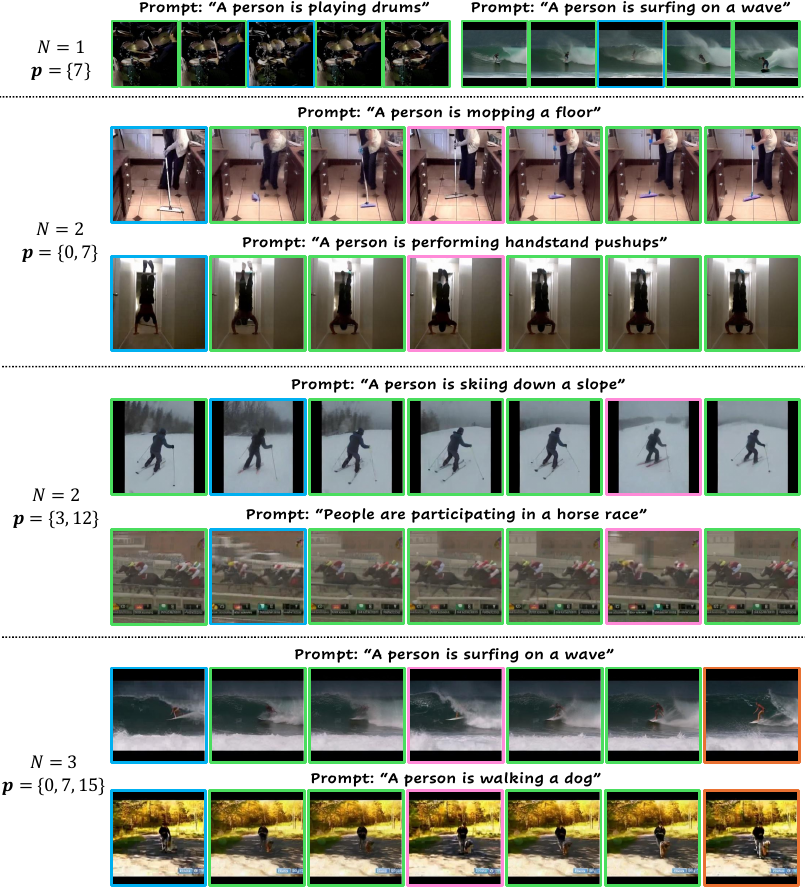}
  \caption{Additional demonstration of videos synthesized by FlexTI2V in new TI2V settings. The images with \textcolor{NavyBlue}{blue}, \textcolor{Lavender}{pink} and \textcolor{orange}{orange} edges are condition imags, and images with \textcolor{ForestGreen}{green} edges are generated video frames. $N$ and $\bm{p}$ are the number and positions of condition images. Our method generates videos with 16 frames in the experiments.}
  \label{fig:additional_demo_new}
\end{figure}

\subsection{Additional Demonstration of Our Method}
\label{sec:additional_demo}

We illustrate more videos synthesized by our method in classic TI2V generation settings (see Fig. \ref{fig:additional_demo_classic}) and new settings (see Fig. \ref{fig:additional_demo_new}). Our approach is able to condition the pretrained T2V diffusion foundation model on any number of images with flexible positions.

\subsection{Failure Cases}
\label{sec:failures}

We demonstrate some failure cases of our method in Fig. \ref{fig:failures} to show the boundary of the capability. We find it is challenging for our method to synthesize smooth camera viewpoint change in output videos (first example). If there is a big difference in viewpoints of two condition images, our method is very likely to treat the two images as irrelevant images and generates a hard transition without a reasonable motion. In addition, we also observe watermarks in a few output videos (second example). This is an inherent bias of the foundation model, which is inherited by our method. Hence, the performance of our method is still limited by the pretrained T2V foundation model.

\section{Implementation Details}
\label{sec:more_implementation}

\begin{table*}[t]
\centering
\small
\setlength{\tabcolsep}{0.1cm}
\begin{tabular}{ll}
\toprule
Class & Prompt \\
\midrule
ApplyEyeMakeup &``A person is applying eye makeup'' \\
ApplyLipstick & ``A person is applying lip stick'' \\
BabyCrawling & ``A baby is crawling'' \\
BenchPress & ``A person is doing a bench press exercise'' \\
Biking & ``A person is riding a bicycle'' \\
BlowDryHair & ``A person is blow drying their hair'' \\
BodyWeightSquats & ``A person is performing bodyweight squats'' \\
BoxingPunchingBag & ``A person is punching a boxing bag'' \\
BrushingTeeth & ``A person is brushing their teeth'' \\
CuttingInKitchen & ``A person is cutting food in the kitchen'' \\
Drumming & ``A person is playing drums'' \\
Fencing & ``Two people are participating in fencing'' \\
Haircut & ``A person is giving someone a haircut'' \\
HandstandPushups & ``A person is performing handstand pushups'' \\
HeadMassage & ``A person is doing a head massage for another one'' \\
HorseRace & ``People are participating in a horse race'' \\
HorseRiding & ``A person is riding a horse'' \\
JumpRope & ``A person is jumping rope'' \\
Kayaking & ``A person is kayaking in water'' \\
Knitting & ``A person is knitting'' \\
Mixing & ``A person is mixing ingredients'' \\
MoppingFloor & ``A person is mopping a floor'' \\
PlayingCello & ``A person is playing a cello'' \\
PlayingFlute & ``A person is playing a flute'' \\
PlayingGuitar & ``A person is playing a guitar'' \\
PlayingPiano & ``A person is playing a piano'' \\
PlayingViolin & ``A person is playing a violin'' \\
PullUps & ``A person is doing pull-ups'' \\
PushUps & ``A person is performing push-ups'' \\
Rafting & ``People are rafting in a river'' \\
RockClimbingIndoor & ``A person is climbing an indoor rock wall'' \\
Rowing & ``People are rowing a boat'' \\
ShavingBeard & ``A person is shaving their beard'' \\
SkateBoarding & ``A person is skateboarding'' \\
Skiing & ``A person is skiing down a slope'' \\
Skijet & ``A person is riding a jet ski'' \\
SkyDiving & ``A person is skydiving'' \\
Surfing & ``A person is surfing on a wave'' \\
TrampolineJumping & ``A person is jumping on a trampoline'' \\
Typing & ``A person is typing on a keyboard'' \\
WalkingWithDog & ``A person is walking a dog'' \\
WritingOnBoard & ``A person is writing on a board''\\
\bottomrule
\end{tabular}
\caption{Prompt for each selected action class in UCF-101. More details of the test set is elaborated in Sec. \ref{sec:test_set}.}
\label{tab:prompts}
\end{table*}

\subsection{Establishment of Test Set}
\label{sec:test_set}

Ni et al. \cite{ni2024ti2v} select 10 classes in UCF-101 which are not diverse enough for a reliable evaluation. The textual prompts used in their work are also oversimplified. In this work, we further extend the test set to 42 classes with manually annotated textual prompt for each action. For each class, there are videos of 25 subjects performing the same action. For diversity, we randomly select one video of each subject in each action class, totaling 1050 test data samples. The prompts are listed in Tab. \ref{tab:prompts}.

\subsection{More Details of Our Method}
\label{sec:our_method_details}

In the proposed method, the model also conditions on textual prompts. We adopt classifier-free guidance in the inference with guidance scale as 9.0. We use a deterministic DDIM sampling strategy ($\sigma_t=0$ in main paper Eq. 2) that can accelerate generation by reducing denoising steps \cite{song2020denoising}. We generate videos with $M=16$ frames in all experiments for a fair comparison with previous methods. Note that our method can naturally extend to videos with more frames. Our method is implemented on a single NVIDIA H100 GPU.

\subsection{Implementation of Previous Methods}
\label{sec:implementation_previous_methods}

We implement all baseline models using their publicly released codebases following the default hyperparameter settings. To adapt TI2V-Zero \cite{ni2024ti2v} to rewinding and outpainting tasks, we reverse the order of frame generation for rewinding, and generate the frames before and after the condition image in two directions separately for outpainting. For TRF \cite{feng2024explorative}, we use the setting of $\verb+gym+$ in their code.

\section{Limitation and Future Work}
\label{sec:limitation_future_work}

Though our approach achieves great performance in many experimental settings compared with previous methods, we still observe some limitations as illustrated in Fig. \ref{fig:failures}. First of all, FlexTI2V is still weak in understanding the transition of camera viewpoints between two condition images. The output video looks like a transition of two slides if there is a big viewpoint change. Second, our method is affected by some flaws in the pretrained foundation models (\textit{e.g.}, watermark) because of the training-free nature. The upperbound of our method is constrained by the T2V model we used. Third, FlexTI2V is able to make the condition images as exact video frames at expected positions. Our method can not synthesize videos following only the subjects or background in the condition images (\textit{e.g.}, personalized video generation).

Based on these limitations, we propose several promising research problems that deserve future investigation.

\begin{itemize}
    \item[$\bullet$] Camera motion modeling is an important problem for vivid video generation. We think camera motion between two video frames can be modeled independently and added to existing training-free video generation approaches.

    \item[$\bullet$] More interpretation work is necessary to probe the bias of pretrained video generation models as well as how to separate the bias from the learned unbiased knowledge.

    \item[$\bullet$] More efforts are needed to extend our method to other image types, such as depth and keypoints.
\end{itemize}

\end{document}